\documentclass[conference]{IEEEtran}
\usepackage{amsmath}
\usepackage{amsfonts}
\usepackage{amssymb}
\usepackage{algorithm}
\usepackage{algpseudocode}  % Use algpseudocode for better line numbering

\usepackage{graphicx}
\usepackage{booktabs}
\usepackage{array}

\usepackage[noadjust]{cite}

\usepackage{url}
\usepackage{multirow}
\usepackage{subcaption}
\usepackage{tikz}
\usepackage{pgfplots}
\pgfplotsset{compat=1.17}
\usetikzlibrary{shapes.geometric, arrows.meta, positioning, fit, backgrounds, patterns}
\usetikzlibrary{patterns.meta}
\usepackage{placeins}
\usepackage{float}
\usepackage{fancyhdr}
\fancypagestyle{firstpage}
{
    \fancyhead[L]{\footnotesize © 2026 IEEE.  Personal use of this material is permitted.  Permission from IEEE must be obtained for all other uses, in any current or future media, including reprinting/republishing this material for advertising or promotional purposes, creating new collective works, for resale or redistribution to servers or lists, or reuse of any copyrighted component of this work in other works. This paper is accepted at the 12th International Conference on Computing and Artificial Intelligence (ICCAI) 2026.}
    \fancyhead[R]{}
}

\begin{document}

\thispagestyle{firstpage}
\title{Mix-and-Match Pruning: Globally Guided Layer-Wise Sparsification of DNNs}

\author{
\IEEEauthorblockN{Danial Monachan\IEEEauthorrefmark{1},
Samira Nazari\IEEEauthorrefmark{2},
Mahdi Taheri\IEEEauthorrefmark{1}\IEEEauthorrefmark{3},
Ali Azarpeyvand\IEEEauthorrefmark{2},
\\Milos Krstic\IEEEauthorrefmark{4},
Michael Hübner\IEEEauthorrefmark{1},
Christian Herglotz\IEEEauthorrefmark{1}}
\IEEEauthorblockA{\IEEEauthorrefmark{1}Brandenburg Technical University, Cottbus, Germany}
\IEEEauthorblockA{\IEEEauthorrefmark{2}University of Zanjan, Iran}
\IEEEauthorblockA{\IEEEauthorrefmark{3}Tallinn University of Technology, Estonia}
\IEEEauthorblockA{\IEEEauthorrefmark{4}Leibniz Institute for High Performance Microelectronics, Frankfurt Oder, Germany}
}

\maketitle

\thispagestyle{firstpage}
\begin{abstract}
Deploying deep neural networks (DNNs) on edge devices requires strong compression with minimal accuracy loss. This paper introduces Mix-and-Match Pruning, a globally guided, layer-wise sparsification framework that leverages sensitivity scores and simple architectural rules to generate diverse, high-quality pruning configurations. The framework addresses a key limitation that different layers and architectures respond differently to pruning, making single-strategy approaches suboptimal. Mix-and-Match derives architecture-aware sparsity ranges, e.g., preserving normalization layers while pruning classifiers more aggressively, and systematically samples these ranges to produce ten strategies per sensitivity signal (magnitude, gradient, or their combination). This eliminates repeated pruning runs while offering deployment-ready accuracy--sparsity trade-offs. Experiments on CNNs and Vision Transformers demonstrate Pareto-optimal results, with Mix-and-Match reducing accuracy degradation on Swin-Tiny by 40\% relative to standard single-criterion pruning. These findings show that coordinating existing pruning signals enables more reliable and efficient compressed models than introducing new criteria.

\end{abstract}


% \FloatBarrier
\section{Introduction}
Deep neural networks deliver strong performance while typically requiring 100--200 MB, and vision transformers often exceed 300--500 MB, making uncompressed models impractical for edge deployment. Although compression approaches such as quantization \cite{quant}, knowledge distillation \cite{hinton2015distilling}, and neural architecture search \cite{zoph2017neural} offer partial solutions, pruning has emerged as one of the most effective strategies, leveraging redundancy in overparameterized networks to dramatically reduce parameters with minimal accuracy degradation \cite{prune, hawx}.

Pruning techniques have advanced along several directions. Early methods use simple importance metrics such as magnitude pruning (MAG) \cite{han2015learning} or gradient-based sensitivity as in SNIP \cite{lee2018snip}. More advanced approaches incorporate second-order information, e.g., GraSP \cite{wang2020picking}, which preserves gradient flow via Hessian approximations. Recent work extends pruning to vision transformers through semi-structured schemes (LPViT) \cite{wu2024lpvit} and joint pruning–quantization frameworks (GETA) \cite{qu2025geta}. Despite their differences—from basic magnitude thresholding to complex gradient-flow reasoning—most methods share a key limitation: each produces a single pruned model for a fixed sparsity target. While their underlying sensitivity signals (magnitude, gradient, or their combinations) can be reused, these methods cannot systematically explore the broader accuracy–compression design space.

Existing pruning methods yield only single-point solutions for fixed sparsity targets, requiring full re-execution when deployment constraints change. They also overlook that different architectural components react differently to pruning.
% skip connections in residual networks must be pruned conservatively to maintain gradient flow \cite{li2017pruning}, whereas attention layers in transformers introduce global dependencies that make them highly sensitive to aggressive sparsification \cite{rao2021dynamicvit}. 
As a result, current single-strategy approaches cannot systematically explore the broader accuracy--compression design space. While recent efforts attempt to generate multiple pruned configurations \cite{desai2024llamaflex}, a unified framework that handles hybrid architectures and provides structured exploration of sparsity trade-offs is still lacking.

The proposed Mix-and-Match Pruning framework addresses this gap through three phases. First, it derives architecture-aware sparsity ranges for each layer using sensitivity scores. Second, it samples these ranges to generate diverse sparsity configurations that span different accuracy--compression trade-offs. Third, it prunes weights using the same sensitivity scores under each configuration. This process yields multiple deployment-ready operating points without re-running the pruning pipeline, enabling substantial and flexible compression for edge scenarios.

The main contributions of this paper are:
\begin{itemize}
    \item A sensitivity-driven pruning framework that establishes architecture-aware sparsity ranges, enabling efficient exploration of the pruning design space without repeated baseline runs.
    \item A systematic strategy-generation mechanism that produces ten pruned models per sensitivity signal, covering a broad set of accuracy--sparsity trade-offs from a single pruning execution.
    \item A comprehensive evaluation on CNNs and ViTs, demonstrating competitive or superior performance compared to classical baselines..
\end{itemize}

The remainder of the paper is organized as follows: Section~\ref{sec:method} details the methodology, Section~\ref{sec:results} reports experimental findings, and Section~\ref{sec:conclusion} concludes the paper.

% \FloatBarrier
\section{Mix-and-Match Pruning Methodology}
\label{sec:method}

\begin{figure}[hbtp]
    \centering
    \includegraphics[width=0.5\textwidth]{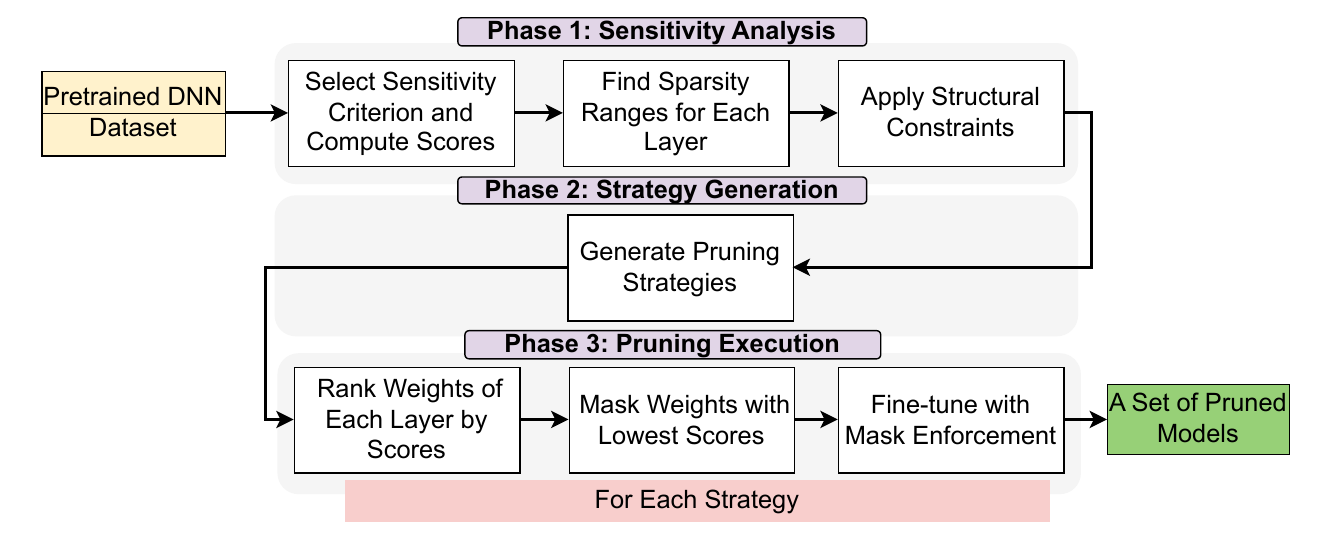}
    \caption{Mix-and-Match Pruning Methodology}
    \label{method}
\end{figure}

Mix-and-Match Pruning generates ten pruned models from a single pruning run through three phases: sensitivity analysis, strategy construction, and pruning with fine-tuning. A ``strategy'' denotes a full layer-wise sparsity vector, not a pruning criterion. Inputs to the framework include a pretrained network, a chosen sensitivity criterion (magnitude, gradient, or magnitude--gradient product), and a small calibration set (5--10\% of training data). The output is a family of pruned models that explore diverse accuracy--sparsity trade-offs. Figure~\ref{method} illustrates the workflow.

\subsection{Phase 1: Sensitivity Analysis and Range Assignment}

Phase~1 corresponds to lines 1--4 of Algorithm~\ref{alg:multistrategy}.  
Sensitivity scores $S_i^{(l)}$ are computed for all prunable weights using one of three criteria:
\begin{equation}
S_i^{(l)} = |w_i^{(l)}|,
\end{equation}
\begin{equation}
S_i^{(l)} = \left|\frac{\partial \mathcal{L}}{\partial w_i^{(l)}}\right|,
\end{equation}
\begin{equation}
S_i^{(l)} = |w_i^{(l)}|\left|\frac{\partial \mathcal{L}}{\partial w_i^{(l)}}\right|.
\end{equation}

These scores determine \emph{which} weights are removed later in Phase~3 but do not specify \emph{how much} pruning each layer should receive.

Each layer is then assigned an architecture-aware pruning range $[\rho_{\min}^{(l)}, \rho_{\max}^{(l)}]$ that constrains its allowable sparsity. These ranges are derived from structural properties that influence pruning tolerance, as well as empirical observations obtained by monitoring multiple existing pruning methods. Layers responsible for maintaining statistical stability, such as normalization layers, generally require minimal or no pruning. Small layers with limited parameters are given narrow ranges to avoid capacity bottlenecks. Fully connected layers, particularly those near the output, tend to contain substantial redundancy and therefore admit more aggressive sparsification. Convolutional layers typically exhibit depth-dependent behaviour, with early layers being more sensitive than deeper layers. Transformer components such as attention projections and MLP blocks also show distinct sensitivity patterns due to their differing functional roles.

These principles provide a unified mechanism for assigning layer-wise sparsity limits across diverse architectures. The specific numerical ranges used in experiments are reported and analysed in Section~\ref{sec:results}, where their empirical validity and consistency are evaluated.

\subsection{Phase 2: Strategy Construction}

Phase~2 corresponds to lines 5--9 of Algorithm~\ref{alg:multistrategy}.  
Given the per-layer sparsity ranges, ten pruning strategies are constructed by sampling a full sparsity vector $\boldsymbol{\rho}_s = [\rho_s^{(1)},\ldots,\rho_s^{(L)}]$ for each strategy $s$. The resulting set spans conservative, aggressive, intermediate, size-aware, and structure-aware behaviours.

Three core strategies assign the minimum, maximum, and midpoint sparsity values across layers. Four additional strategies interpolate within each layer's range using coefficients $\alpha \in \{0.3, 0.5, 0.7, 0.9\}$ via
\[
\rho^{(l)} = \rho_{\min}^{(l)} + \alpha\big(\rho_{\max}^{(l)} - \rho_{\min}^{(l)}\big),
\]
providing progressively more aggressive pruning while remaining within allowable bounds.

A parameter-proportional strategy scales sparsity according to the relative size of each layer:
\[
\rho^{(l)} = \rho_{\min}^{(l)} + \beta\big(\rho_{\max}^{(l)} - \rho_{\min}^{(l)}\big), \qquad
\beta = \min\!\left(1,\; \frac{|\mathcal{W}^{(l)}|}{|\mathcal{W}_{\text{avg}}|} \cdot 0.1\right),
\]
where $|\mathcal{W}^{(l)}|$ denotes the number of weights in layer $l$ and $|\mathcal{W}_{\text{avg}}|$ is the network-wide average. This biases sparsity upward in large layers while avoiding excessive pruning in small ones.

Finally, two structure-aware strategies adjust the interpolation coefficient according to layer depth and functional role, yielding classifier-heavy and feature-heavy variants that modify sparsity distribution without exceeding the assigned per-layer ranges.

\subsection{Phase 3: Pruning and Fine-Tuning}

Phase~3 corresponds to lines 10--17 of Algorithm~\ref{alg:multistrategy}.  
For each strategy, the base model is copied, and pruning is performed layer-wise by ranking weights using $S_i^{(l)}$ and removing the lowest $\rho_s^{(l)}$ fraction via binary masks $\mathcal{M}_s^{(l)}$. Masked weights are set to zero and remain fixed during fine-tuning:

\[
\tilde{w}_{s,i}^{(l)} = \mathcal{M}_{s,i}^{(l)} \cdot w_i^{(l)}.
\]

Fine-tuning is then applied with mask enforcement after each update to ensure pruned weights stay zeroed. This produces ten pruned models per sensitivity criterion, each reflecting a different layer-wise sparsity distribution but sharing the same underlying importance scores.

\begin{algorithm}
\caption{Mix-and-Match Pruning Framework}
\label{alg:multistrategy}
\begin{algorithmic}[1]

\Require Pretrained network $\mathcal{N}$, criterion $c$, calibration data $\mathcal{D}_{cal}$
\Ensure Pruned models $\{\mathcal{N}_{s}\}_{s=1}^{10}$

\Statex \textit{--- Phase 1: Sensitivity Analysis ---}
\State Compute scores $S_i^{(l)}$ using $c$ on $\mathcal{D}_{cal}$
\For{each layer $l$}
    \State Identify layer type and set pruning range $[\rho_{\min}^{(l)},\rho_{\max}^{(l)}]$
\EndFor

\Statex \textit{--- Phase 2: Strategy Construction ---}
\State $\mathcal{S} \gets \varnothing$
\For{each strategy $s$}
    \State Sample pruning vector $\boldsymbol{\rho}_s$
    \State $\mathcal{S} \gets \mathcal{S} \cup \{\boldsymbol{\rho}_s\}$
\EndFor

\Statex \textit{--- Phase 3: Pruning \& Fine-tuning ---}
\For{each $\boldsymbol{\rho}_s \in \mathcal{S}$}
    \State Copy $\mathcal{N}$ to $\mathcal{N}_s$
    \For{each layer $l$}
        \State Sort weights by $S_i^{(l)}$ and prune lowest $\rho_s^{(l)}$ to form mask $\mathcal{M}_s^{(l)}$
        \State Apply mask: $w_i^{(l)} \gets w_i^{(l)} \mathcal{M}_{s,i}^{(l)}$
    \EndFor
    \State Fine-tune $\mathcal{N}_s$ with masks enforced
\EndFor

\Return $\{\mathcal{N}_1,\ldots,\mathcal{N}_{10}\}$

\end{algorithmic}
\end{algorithm}

\subsection{Computational Complexity Analysis}
\label{sec:complexity}

The Mix-and-Match framework introduces a lightweight one-time setup cost in Phase~1 (sensitivity-driven range establishment) that is amortized across the generation of multiple pruning strategies. This section analyzes the computational overhead compared to baseline pruning methods.

\textbf{Complexity of Phase 1 (Sensitivity-Driven Range Establishment).}
Phase~1 computes sensitivity scores once using the user-selected criterion (magnitude, gradient, or product) on a small calibration set, then applies architecture-aware rules to establish layer-wise sparsity ranges. Let $N$ denote the total number of prunable weights across all $L$ layers. The time complexity is:

\begin{equation}
\mathcal{O}_{\text{Phase 1}} = \mathcal{C}_{\text{sensitivity}}(N) + \mathcal{O}(L),
\end{equation}

where $\mathcal{C}_{\text{sensitivity}}(N)$ depends on the chosen criterion: $\mathcal{O}(N)$ for magnitude-based scoring (weight access), or $\mathcal{O}(N \times B_{\text{cal}})$ for gradient-based scoring (forward-backward pass over calibration batches $B_{\text{cal}}$). The architecture-aware rule application is $\mathcal{O}(L)$ (linear in the number of layers). Critically, this computation happens \emph{once}, not repeatedly, making Phase~1 significantly more efficient than iterative baseline execution approaches.

\textbf{Complexity of Phase 2 (Strategy Generation).}
Phase~2 is negligible, consisting only of interpolation operations within pre-established ranges. Generating all 10 strategies requires $\mathcal{O}(L)$ arithmetic operations and is completed in milliseconds.

\textbf{Complexity of Phase 3 (Pruning Execution).}
Each of the 10 strategies reuses the pre-computed sensitivity scores from Phase~1 to rank and mask weights, followed by fine-tuning. The user selects one sensitivity criterion for the entire run. The cost is:

\begin{equation}
\mathcal{O}_{\text{Phase 3}} = 10 \times \mathcal{C}_{\text{mask+ft}}(N),
\end{equation}

where $\mathcal{C}_{\text{mask+ft}}(N)$ includes weight ranking, binary mask application ($\mathcal{O}(N \log N)$ for sorting), and fine-tuning. Since fine-tuning dominates this cost ($\mathcal{O}(E \times N \times B)$, where $E$ denotes the number of fine-tuning epochs and $B$ the number of mini-batches per epoch) and is identical across all methods, the per-model overhead is equivalent to that of a single baseline execution.

% \textbf{Comparison with Baselines.}
Running a single baseline method (e.g., MAG) at one target sparsity requires $\mathcal{C}_{\text{sensitivity}}(N) + \mathcal{C}_{\text{mask+ft}}(N)$ operations. To obtain the same diversity of solutions using traditional baseline pruning methods, one would need to: (1) run multiple baselines (MAG, SNIP, GraSP) separately at various global sparsity targets, (2) manually tune per-layer sparsity configurations through trial-and-error, and (3) re-execute the entire pruning pipeline for each new configuration. The total cost for generating 10 comparable configurations with baselines would be at least $10 \times [\mathcal{C}_{\text{sensitivity}}(N) + \mathcal{C}_{\text{mask+ft}}(N)]$, plus significant hyperparameter search overhead.

% \textbf{Efficiency Advantage.}
Mix-and-Match computes sensitivity scores once in Phase~1, then reuses them across all 10 strategies in Phase~3. The total cost is:

\begin{equation}
\mathcal{C}_{\text{total}} = \mathcal{C}_{\text{sensitivity}}(N) + 10 \times \mathcal{C}_{\text{mask+ft}}(N).
\end{equation}

In contrast, generating 10 diverse pruned models with baseline methods requires:

\begin{equation}
\mathcal{C}_{\text{baseline}} = 10 \times [\mathcal{C}_{\text{sensitivity}}(N) + \mathcal{C}_{\text{mask+ft}}(N)].
\end{equation}

Since fine-tuning dominates the cost ($\mathcal{C}_{\text{mask+ft}}(N) \gg \mathcal{C}_{\text{sensitivity}}(N)$), the framework's overhead is marginal. Crucially, Mix-and-Match provides: (1) multi-point exploration—10 strategies spanning the accuracy-sparsity trade-off space from a single sensitivity computation; (2) architecture-aware bounds—empirically validated ranges prevent over- or under-pruning; and (3) systematic coverage—strategies are generated deterministically, ensuring reproducibility without manual hyperparameter tuning. For scenarios requiring multiple deployment configurations or exploration of the Pareto front, the framework is strictly more efficient than repeated baseline execution.

% Algorithm~\ref{alg:multistrategy} presents the complete framework workflow. Lines 5-12 (Phase 1) establish architecture-aware ranges by executing baseline methods at multiple global sparsity targets and computing per-layer minimum and maximum observed sparsities. Lines 15-22 (Phase 2) generate the strategy family through structured sampling of established ranges, producing ten diverse pruning configurations. Lines 24-33 (Phase 3) execute each strategy with sensitivity scores computed once and cached (line 25) to avoid redundant calculations across all ten strategies, followed by mask application, fine-tuning with gradient updates restricted to unpruned weights, and post-training quantization. The framework outputs a family of pruned models spanning the accuracy-compression tradeoff space, enabling deployment flexibility without re-execution.

% \textbf{Computational Considerations.} The range establishment phase (lines 5-12) requires running three baseline methods at multiple sparsity targets, representing a one-time setup cost of approximately three times a single baseline evaluation. However, this cost is amortized across the ten generated strategies and can be reused for future pruning tasks on the same architecture. Importantly, the reported inference FLOPs measure per-forward-pass computational requirements of the final pruned models and are identical across methods because unstructured pruning maintains layer dimensions. Training and setup costs are one-time expenses not included in deployment metrics, following standard practice in pruning literature.

\section{Experimental Evaluation}
\label{sec:results}
\subsection{Experimental Setup}

% \vspace{-0.2cm}

The evaluation covers four representative architectures selected to span classical convolutional, residual, hybrid, and hierarchical transformer designs. These models, including VGG-11, ResNet-18, LeViT-384, and Swin-Tiny, are paired with datasets commonly used for vision benchmarks. VGG-11 and LeViT-384 are trained on CIFAR-10, ResNet-18 on GTSRB, and Swin-Tiny on CIFAR-100. The baseline characteristics of all models, including parameter counts and top-1 accuracies, are summarized in Table~\ref{tab:baselines}. These baselines provide the reference point against which the effects of the Mix-and-Match Pruning are assessed.

\begin{table}[htbp]
\centering
\captionsetup{justification=centering}
\caption{Baseline model characteristics.}
\label{tab:baselines}
\begin{tabular}{lccc}
\toprule
Model & Dataset & Parameters & Accuracy (\%) \\
\midrule
VGG-11 & CIFAR-10 & 28.1M & 92.40 \\
ResNet-18 & GTSRB & 11.2M & 99.32 \\
LeViT-384 & CIFAR-10 & 37.6M & 97.75 \\
Swin-Tiny & CIFAR-100 & 27.6M & 89.14 \\
\bottomrule
\end{tabular}
\end{table}

Since unstructured pruning maintains original tensor dimensions, theoretical Floating Point Operations (FLOPs) remain unchanged. Nonetheless, sparsification substantially reduces the memory footprint of the resulting models and enables acceleration on sparse-aware hardware. These properties make the baselines in Table~\ref{tab:baselines} suitable for examining how globally guided, layer-adaptive sparsity interacts with architectural diversity and dataset scale.

All pruned models are fine-tuned to recover accuracy under a unified optimization protocol to ensure comparability across architectures and pruning strategies. The Adam optimizer is used for VGG-11, ResNet-18, and Swin-Tiny with learning rates of $1 \times 10^{-4}$ (VGG-11) and $5 \times 10^{-5}$ (ResNet-18, Swin-Tiny). For LeViT-384, AdamW~\cite{loshchilov2019decoupled} with weight decay 0.05 and learning rate $5 \times 10^{-5}$ is employed due to its hybrid architecture. All experiments use a batch size of 128. During fine-tuning, parameter masks are enforced after every gradient update to preserve the imposed sparsity pattern. Early stopping with a patience of 3 epochs is employed to prevent overfitting. The number of fine-tuning epochs is capped at 20 for VGG-11, ResNet-18, and LeViT-384, while Swin-Tiny uses up to 30 epochs due to its hierarchical window-based attention, which demonstrates slower convergence when subjected to high sparsity.

Five representative baseline pruning methods are included: MAG (magnitude pruning)~\cite{han2015learning}, which removes weights with the smallest absolute values; SNIP (Single-shot Network Pruning)~\cite{lee2018snip}, which uses connection sensitivity based on the product of gradients and weights before training; GraSP (Gradient Signal Preservation)~\cite{wang2020picking}, which leverages gradient flow and Hessian information to identify and preserve critical connections; LPViT~\cite{wu2024lpvit}, which applies semi-structured pruning optimized for vision transformer architectures; and GETA~\cite{qu2025geta}, which employs joint optimization with quantization-aware importance scoring. All baseline methods follow identical fine-tuning protocols with the same learning rates, batch sizes, and epoch limits to ensure fair comparison.

Architecture-aware structural constraints are incorporated into Mix-and-Match Pruning to prevent degradation from overly aggressive sparsification in sensitive components. The specific thresholds used in all experiments are:

\begin{itemize}
    \item \textbf{Normalization layers} (BatchNorm, LayerNorm): fixed at $0\%$ sparsity (unprunable) to preserve statistical stability~\cite{li2017pruning}.
    \item \textbf{Small layers} with fewer than 10K parameters: capped at a conservative range of $[0\%, 10\%]$, as empirical analysis shows that aggressive pruning disproportionately harms accuracy for such layers~\cite{han2015learning}.
    \item \textbf{Transformer patch embeddings}: restricted to $[15\%, 30\%]$ sparsity to maintain token diversity and avoid collapse of early-stage representations.
\end{itemize}

These empirically derived ranges guide feasible sparsity exploration and are further analyzed and justified in the results.

The evaluation relies on several metrics that capture sparsification level and predictive performance. Sparsity is defined as the global percentage of weights set to zero across all prunable layers, computed as the ratio of zero-valued weights to the total number of weights. Bias parameters are excluded, and normalization layers (BatchNorm, LayerNorm) are preserved through architecture-aware constraints. Predictive performance is measured using FP32 accuracy after fine-tuning. The accuracy drop is expressed as the difference between the unpruned FP32 baseline and the pruned model.

% \FloatBarrier

% \FloatBarrier

\begin{table*}
\centering
\captionsetup{justification=centering}
\caption{Mix-and-Match Pruning Results Across All Architectures.}
\label{tab:summary_all}
\resizebox{0.9\textwidth}{!}{%
% \scriptsize
% \setlength{\tabcolsep}{3pt}
\begin{tabular}{l|ccc|ccc|ccc|ccc}
\toprule
\multirow{2}{*}{\centering\arraybackslash\textbf{Strategy}}
& \multicolumn{3}{c|}{\textbf{VGG-11}}
& \multicolumn{3}{c|}{\textbf{ResNet-18}}
& \multicolumn{3}{c|}{\textbf{LeViT-384}}
& \multicolumn{3}{c}{\textbf{Swin-Tiny}} \\
\cmidrule(lr){2-4} \cmidrule(lr){5-7} \cmidrule(lr){8-10} \cmidrule(lr){11-13}
& \multicolumn{1}{c}{Spar.} & \multicolumn{1}{c}{Acc.} & \multicolumn{1}{c|}{Drop}
& \multicolumn{1}{c}{Spar.} & \multicolumn{1}{c}{Acc.} & \multicolumn{1}{c|}{Drop}
& \multicolumn{1}{c}{Spar.} & \multicolumn{1}{c}{Acc.} & \multicolumn{1}{c|}{Drop}
& \multicolumn{1}{c}{Spar.} & \multicolumn{1}{c}{Acc.} & \multicolumn{1}{c|}{Drop} \\
& \multicolumn{1}{c}{(\%)} & \multicolumn{1}{c}{(\%)} & \multicolumn{1}{c|}{(\%)}
& \multicolumn{1}{c}{(\%)} & \multicolumn{1}{c}{(\%)} & \multicolumn{1}{c|}{(\%)}
& \multicolumn{1}{c}{(\%)} & \multicolumn{1}{c}{(\%)} & \multicolumn{1}{c|}{(\%)}
& \multicolumn{1}{c}{(\%)} & \multicolumn{1}{c}{(\%)} & \multicolumn{1}{c|}{(\%)} \\
\midrule
Max-Aggressive      & \textbf{90.40} & \textbf{90.25} & \textbf{2.15} & 89.61 & 97.85 & 1.48 & \textbf{64.07} & \textbf{97.49} & \textbf{0.26} & \textbf{49.75} & \textbf{87.30} & \textbf{2.38} \\
Min-Conservative    & 89.20 & 90.20 & 2.14 & 72.78 & 97.51 & 1.80 & 44.09 & 97.45 & 0.30 & 33.19 & 86.72 & 2.67 \\
Balanced            & 89.60 & 89.85 & 2.51 & 82.79 & 97.19 & 2.14 & 54.08 & 97.54 & 0.21 & 41.47 & 87.65 & 1.71 \\
Classifier-Heavy & 89.50 & 90.21 & 2.17 & -- & -- & -- & 44.10 & 97.56 & 0.19 & -- & -- & -- \\
Conv-Heavy      & 88.31 & 90.19 & 2.20 & -- & -- & -- & 59.25 & 97.50 & 0.25 & 42.60 & 87.20 & 2.77 \\
FC-Heavy            & -- & -- & -- & 72.84 & 97.54 & 1.78 & -- & -- & -- & -- & -- & -- \\
Late-Stage-Aggressive    & -- & -- & -- & 87.55 & 97.98 & 1.33 & -- & -- & -- & -- & -- & -- \\
Attention-Aggressive     & -- & -- & -- & -- & -- & -- & -- & -- & -- & 44.61 & 86.90 & 2.66 \\
Lower-30th-Percentile     & 89.68 & 90.10 & 2.27 & 78.79 & 96.85 & 2.47 & 50.08 & 97.54 & 0.21 & 38.16 & 87.08 & 2.79 \\
Middle-50th-Percentile    & 89.60 & 90.41 & 1.97 & 82.79 & 97.21 & 2.11 & 54.08 & 97.41 & 0.34 & 41.47 & 87.17 & 2.39 \\
Upper-70th-Percentile     & 91.52 & 90.23 & 2.17 & 86.80 & 97.19 & 2.17 & 58.07 & 97.54 & 0.21 & 44.78 & 87.64 & 1.68 \\
Upper-90th-Percentile     & 91.84 & 89.96 & 2.38 & \textbf{90.06} & \textbf{97.85} & \textbf{1.47} & 62.07 & 97.33 & 0.42 & 48.10 & 87.20 & 2.75 \\
Parameter-Proportional         & 90.24 & 90.28 & 2.10 & \textbf{87.28} & \textbf{98.11} & \textbf{1.21} & 48.69 & 97.50 & 0.25 & 40.93 & 87.10 & 2.75 \\
\midrule
\textbf{Mean $\pm$ Std}
& 90.09 & 90.17 & 2.19
& 82.95 & 97.63 & 1.80
& 53.86 & 97.49 & 0.25
& 42.21 & 87.20 & 2.42 \\
& $\pm$1.03 & $\pm$0.17 & $\pm$0.15
& $\pm$6.61 & $\pm$0.43 & $\pm$0.39
& $\pm$6.95 & $\pm$0.07 & $\pm$0.06
& $\pm$5.08 & $\pm$0.31 & $\pm$0.43 \\
\bottomrule
\end{tabular}%
}
\end{table*}

% Table III: Comparison of Mix-and-Match vs. Baseline Pruning Methods
% \FloatBarrier

\begin{figure*}[!tb]
    \centering
    \includegraphics[width=\textwidth]{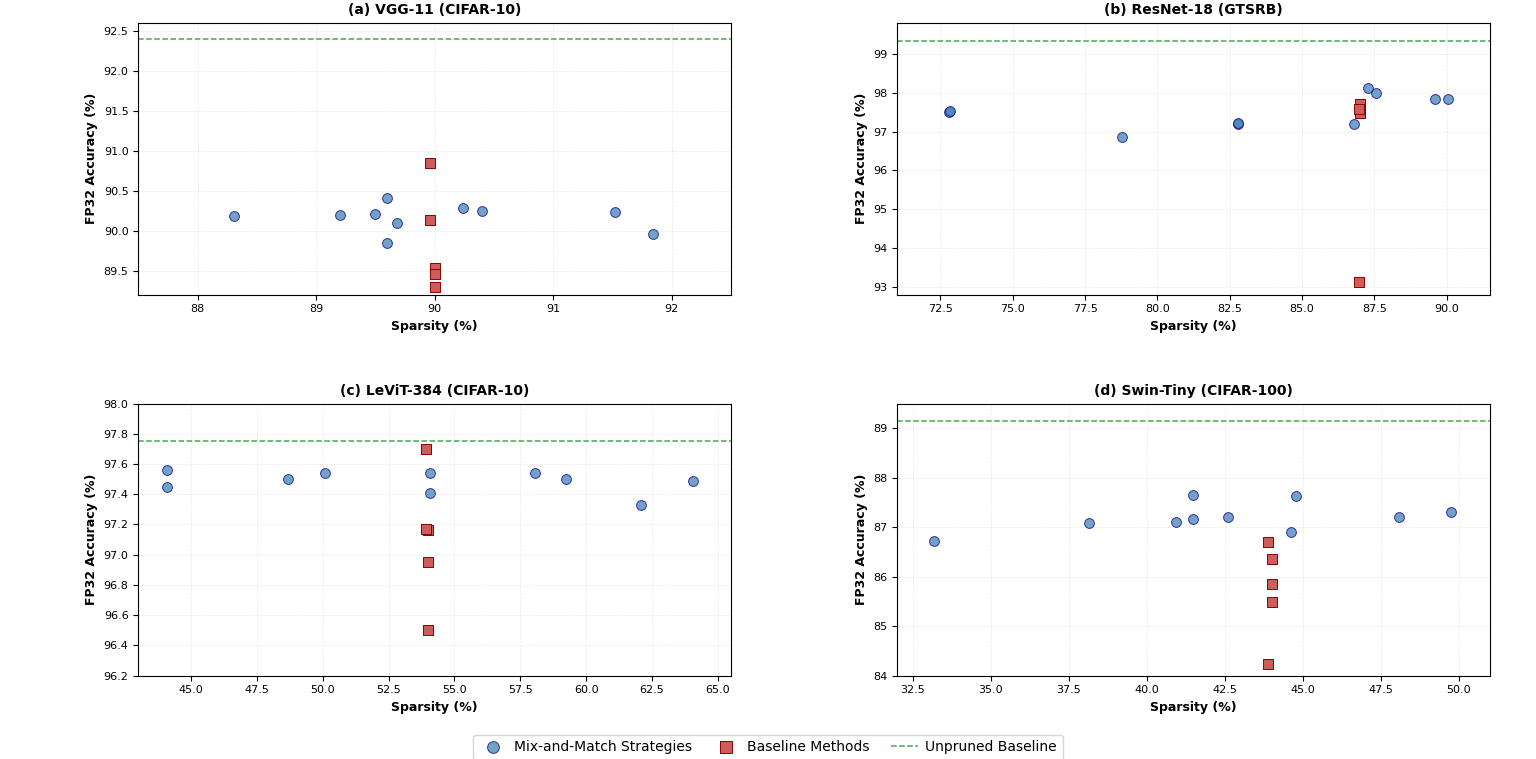}
    \caption{Sparsity vs.\ accuracy trade-offs for Mix-and-Match pruning across four architectures.}
    \label{fig:scatter_all}
\end{figure*}

\begin{table}
    \centering
\captionsetup{justification=centering}
\caption{Comparison of Mix-and-Match vs. Baseline Pruning Methods.}
\label{tab:baseline_comparison}
\begin{subtable}{\columnwidth}
\centering
\captionsetup{justification=centering}
\caption{ CNNs}
\label{tab:baseline_comparison_a}
\resizebox{\columnwidth}{!}{%
\scriptsize
\setlength{\tabcolsep}{6pt}
\begin{tabular}{l|ccc|ccc}
\hline
\multicolumn{1}{c|}{\textbf{Method}} &
\multicolumn{3}{c|}{\textbf{VGG-11 @}\,$\sim$90\%} &
\multicolumn{3}{c}{\textbf{ResNet-18 @}\,$\sim$87\%} \\
\hline
& \multicolumn{1}{c}{Spar.} & \multicolumn{1}{c}{Acc.} & \multicolumn{1}{c|}{Drop}
& \multicolumn{1}{c}{Spar.} & \multicolumn{1}{c}{Acc.} & \multicolumn{1}{c}{Drop} \\
& \multicolumn{1}{c}{(\%)} & \multicolumn{1}{c}{(\%)} & \multicolumn{1}{c|}{(\%)}
& \multicolumn{1}{c}{(\%)} & \multicolumn{1}{c}{(\%)} & \multicolumn{1}{c}{(\%)} \\
Mix-and-Match
& 90.40 & 90.25 & 2.15
& \textbf{87.28} & \textbf{98.11} & \textbf{1.21} \\
LPViT
& \textbf{89.96} & \textbf{90.85} & \textbf{1.55}
& 86.96 & 97.59 & 1.73 \\
GETA
& 89.96 & 90.13 & 2.27
& 86.96 & 93.14 & 6.18 \\
MAG
& 90.00 & 89.53 & 2.87
& 87.00 & 97.55 & 1.77 \\
SNIP
& 90.00 & 89.46 & 2.94
& 87.00 & 97.47 & 1.85 \\
GraSP
& 90.00 & 89.30 & 3.10
& 87.00 & 97.70 & 1.62 \\
\end{tabular}%
}
\end{subtable}

\vspace{0.3cm}

\begin{subtable}{\columnwidth}
\centering
\captionsetup{justification=centering}
\caption{ ViTs}
\label{tab:baseline_comparison_b}
\resizebox{\columnwidth}{!}{%
\scriptsize
\setlength{\tabcolsep}{6pt}
\begin{tabular}{l|ccc|ccc}
\hline
\multicolumn{1}{c|}{\textbf{Method}} &
\multicolumn{3}{c|}{\textbf{LeViT-384 @}\,$\sim$54\%} &
\multicolumn{3}{c}{\textbf{Swin-Tiny @}\,$\sim$45\%} \\
\hline
& \multicolumn{1}{c}{Spar.} & \multicolumn{1}{c}{Acc.} & \multicolumn{1}{c|}{Drop}
& \multicolumn{1}{c}{Spar.} & \multicolumn{1}{c}{Acc.} & \multicolumn{1}{c}{Drop} \\
& \multicolumn{1}{c}{(\%)} & \multicolumn{1}{c}{(\%)} & \multicolumn{1}{c|}{(\%)}
& \multicolumn{1}{c}{(\%)} & \multicolumn{1}{c}{(\%)} & \multicolumn{1}{c}{(\%)} \\
Mix-and-Match
& 54.08 & 97.54 & 0.21
& \textbf{44.78} & \textbf{87.64} & \textbf{1.50} \\
LPViT
& \textbf{53.91} & \textbf{97.70} & \textbf{0.05}
& 43.88 & 86.70 & 2.44 \\
GETA
& 53.91 & 97.17 & 0.58
& 43.88 & 84.23 & 4.91 \\
MAG
& 54.00 & 96.50 & 1.25
& 44.00 & 86.35 & 2.79 \\
SNIP
& 54.00 & 96.95 & 0.80
& 44.00 & 85.85 & 3.29 \\
GraSP
& 54.00 & 97.16 & 0.59
& 44.00 & 85.48 & 3.66 \\
\end{tabular}%
}
\end{subtable}
\end{table}

\textbf{Practical Runtime.}
On an NVIDIA A100 GPU, Phase~1 (sensitivity computation + rule application) for VGG-11 completes in under 2 minutes, while each Phase~3 strategy execution (pruning + fine-tuning for 20 epochs) requires 8--10 minutes. Generating all 10 strategies takes roughly 85 minutes total (2 min + 10$\times$8.3 min), compared to approximately 100 minutes for 10 separate baseline executions (10$\times$10 min), with the added benefit of systematic exploration and architecture-aware sparsity allocation.

% \FloatBarrier
\subsection{Pruning Results}

The Mix-and-Match framework addresses a multi-objective optimization problem with two competing goals: maximizing sparsity (compression) while maximizing accuracy (performance). In this context, a strategy is considered \textit{Pareto-efficient} (or non-dominated) if no other strategy achieves both higher accuracy and higher sparsity simultaneously. The ``best-performing'' strategy for a given architecture refers to the solution with the lowest accuracy drop among configurations at comparable sparsity levels, representing a favorable point on the Pareto front. Table~\ref{tab:summary_all} presents comprehensive results across all strategies for all four architectures, with strategies spanning diverse trade-off points to enable flexible deployment. Table~\ref{tab:baseline_comparison} demonstrates how Mix-and-Match compares against established baseline methods at matched sparsity levels.

The optimal sensitivity criterion varies by architecture: VGG-11 benefits from the product of magnitude and gradient, yielding a mean FP32 accuracy of 90.17\% ± 0.17\% at approximately 90\% sparsity with 2.19\% ± 0.15\% drop. ResNet-18 uses magnitude-only sensitivity, achieving a mean FP32 accuracy of 97.63\% ± 0.43\% at approximately 83\% sparsity with 1.80\% ± 0.39\% drop. LeViT-384 uses gradient-only sensitivity, achieving a mean FP32 accuracy of 97.49\% ± 0.07\% at approximately 54\% sparsity with only 0.25\% ± 0.06\% drop. Swin-Tiny uses magnitude-only sensitivity, achieving a mean FP32 accuracy of 87.20\% ± 0.31\% at approximately 42\% sparsity with 2.42\% ± 0.43\% drop.
% Table~\ref{tab:swin_baselines} provides a comparison with established pruning methods at matched sparsity.

% \FloatBarrier

Table~\ref{tab:summary_all} provides a comprehensive side-by-side comparison of all pruning strategies across the four evaluated architectures. The Mix-and-Match framework generates diverse Pareto-optimal configurations: VGG-11 Max-Aggressive achieves 90.40\% sparsity with 90.25\% accuracy and 2.15\% drop, ResNet-18 reaches 90.06\% sparsity with 97.85\% accuracy and 1.47\% drop (Upper-90th-Percentile) or 87.28\% sparsity with 98.11\% accuracy and 1.21\% drop (Parameter-Proportional), LeViT-384 Max-Aggressive attains 64.07\% sparsity with 97.49\% accuracy and only 0.26\% drop, and Swin-Tiny Max-Aggressive achieves 49.75\% sparsity with 87.30\% accuracy and 2.38\% drop. When compared with baseline methods in Table~\ref{tab:baseline_comparison}, Mix-and-Match achieves the best performance on ResNet-18 and Swin-Tiny, while LPViT demonstrates superior results on VGG-11 and LeViT-384, highlighting complementary strengths across different architectures. Notably, the mean accuracy standard deviations are exceptionally low (VGG-11: $\pm$0.17\%, ResNet-18: $\pm$0.43\%, LeViT-384: $\pm$0.07\%, Swin-Tiny: $\pm$0.31\%), demonstrating the framework's robustness and consistency across diverse strategy configurations (Figure \ref{fig:scatter_all}.

The experimental results demonstrate that the proposed Mix-and-Match Pruning framework achieves competitive or superior performance compared to baseline methods across all evaluated architectures, including VGG-11, ResNet-18, LeViT-384, and Swin-Tiny. By combining layer-wise sensitivity exploration with adaptive sparsity allocation, the framework identifies pruning configurations that maintain high accuracy even at aggressive sparsity levels.

Layer-wise pruning allows the framework to assign different sparsity levels to each layer based on its sensitivity, preserving crucial features in early convolutional layers, attention heads, or residual connections while aggressively pruning less critical regions. This approach results in higher overall accuracy compared to uniform or single-criterion pruning.  

The results indicate that the optimal sensitivity metric depends on the architecture: a combination of magnitude and gradient is most effective for VGG-11, magnitude alone for ResNet-18 and Swin-Tiny, and gradient alone for LeViT-384. This variation reflects architectural properties: VGG-11's sequential structure benefits from combining learned feature importance (magnitude) with training dynamics (gradient); ResNet-18 and Swin-Tiny's residual connections and hierarchical attention ensure stable gradient flow, making magnitude a reliable indicator of parameter importance; while LeViT-384's hybrid architecture with dynamic attention routing requires gradient-based sensitivity to capture context-dependent parameter utility during training.

ResNet-18 exhibits two distinct Pareto-optimal configurations: Upper-90th-Percentile achieves maximum sparsity (90.06\%) with 97.85\% accuracy, while Parameter-Proportional offers a conservative trade-off (87.28\% sparsity, 98.11\% accuracy, only 1.21\% drop), demonstrating the framework's ability to produce multiple deployment-appropriate solutions tailored to different resource constraints.

\textbf{Architecture-Specific Observations.} While Mix-and-Match achieves competitive results across all architectures, Swin-Tiny exhibits more modest compression rates (49.75\% sparsity) compared to VGG-11 (90.40\%) and ResNet-18 (90.06\%). This reflects the inherent characteristics of hierarchical Vision Transformers: window-based attention mechanisms create strong inter-layer dependencies that limit layer-wise pruning effectiveness. The framework identifies stable operating points for Swin-Tiny while maintaining accuracy within 2.38\% of the baseline, demonstrating that architecture-aware pruning successfully adapts to structural constraints even when aggressive compression is limited by architectural design.

\section{Conclusion}
\label{sec:conclusion}

This paper presented \textit{Mix-and-Match Pruning}, an architecture-aware framework that improves standard pruning methods through layer-wise sensitivity analysis and adaptive sparsity allocation. The approach requires no new pruning rules or retraining, yet consistently reduces accuracy degradation across CNNs and Vision Transformers. The framework enables strong compression while maintaining accuracy, supporting deployment on memory-limited devices, shrinking VGG-11 to $\sim$10 MB and ResNet-18 to $\sim$4.5 MB.\footnote{Memory footprints computed as parameter count $\times$ 4 bytes (FP32).} A key insight is that architectural structure determines compressibility. Mix-and-Match incorporates this through architecture-aware sparsity ranges, automatically generating multiple viable operating points without manual tuning. On Swin-Tiny, the method reduces accuracy degradation by 40\% relative to standard single-criterion pruning, demonstrating its practical advantage for edge deployment. Future work will extend the framework to structured pruning patterns, larger datasets, and emerging architectures.

\section*{Acknowledgements}
\scriptsize 
This work was supported in part by the Estonian Research Council grant PUT PRG1467 "CRASHLESS“, EU Grant Project 101160182 “TAICHIP“, by the Deutsche Forschungsgemeinschaft (DFG, German Research Foundation) – Project-ID "458578717", and by the Federal Ministry of Research, Technology and Space of Germany (BMFTR) for supporting Edge-Cloud AI for DIstributed Sensing and COmputing (AI-DISCO) project (Project-ID "16ME1127").

\bibliographystyle{IEEEtran}
\bibliography{reference}

\end{document}